\documentclass{article}

\usepackage{PRIMEarxiv}

\usepackage[utf8]{inputenc} 
\usepackage[T1]{fontenc}    
\usepackage{hyperref}       
\usepackage{url}            
\usepackage{booktabs}       
\usepackage{amsfonts}       
\usepackage{nicefrac}       
\usepackage{microtype}      
\usepackage{lipsum}
\usepackage{fancyhdr}       
\usepackage{graphicx}       
\graphicspath{{media/}}     

\usepackage{xcolor}         
\usepackage{algorithm}
\usepackage{algorithmic}
\usepackage{amsmath}
\usepackage{amssymb}
\usepackage{mathtools}
\usepackage{wrapfig}
\usepackage{enumitem}
\usepackage{minitoc}
\newcommand{\model}{\mbox{\textsc{Clarifier}}}
\newcommand{\suggest}{\mbox{\textsc{Suggest}}}

\title{Beyond Active Learning: Leveraging the Full Potential of Human Interaction via Auto-Labeling, Human Correction, and Human Verification} 

\author{%
    Nathan Beck\\
    University of Texas, Dallas\\
    nathan.beck@utdallas.edu
    \And
    Krishnateja Killamsetty\\
    University of Texas, Dallas\\
    krishnateja.killamsetty@utdallas.edu
    \And
    Suraj Kothawade\\
    University of Texas, Dallas\\
    suraj.kothawade@utdallas.edu
    \And
    Rishabh Iyer\\
    University of Texas, Dallas\\
    rishabh.iyer@utdallas.edu
}

\begin{document}
\doparttoc
\faketableofcontents
\maketitle

\begin{abstract}
    Active Learning (AL) is a human-in-the-loop framework to interactively and adaptively label data instances, thereby enabling significant gains in model performance compared to random sampling. AL approaches function by selecting the hardest instances to label, often relying on notions of diversity and uncertainty. However, we believe that these current paradigms of AL do not leverage the full potential of human interaction granted by automated label suggestions. Indeed, we show that for many classification tasks and datasets, most people verifying if an automatically suggested label is correct take $3\times$ to $4\times$ less time than they do changing an incorrect suggestion to the correct label (or labeling from scratch without any suggestion). Utilizing this result, we propose \model\ (aCtive LeARnIng From tIEred haRdness), an Interactive Learning framework that admits more effective use of human interaction by leveraging the reduced cost of verification. By targeting the hard (uncertain) instances with existing AL methods, the intermediate instances with a novel label suggestion scheme using submodular mutual information functions on a per-class basis, and the easy (confident) instances with highest-confidence auto-labeling, \model\ can improve over the performance of existing AL approaches on multiple datasets -- particularly on those that have a large number of classes -- by almost 1.5$\times$ to 2$\times$ in terms of relative labeling cost.
\end{abstract}

\section{Introduction}
\label{sec:introduction}

In recent years, machine learning practitioners have employed the use of deep neural networks (DNNs) to achieve state-of-the-art performance on many tasks, such as image classification, object detection, language translation, and so forth. The performance of these DNNs largely owe their success to increasingly large datasets used in training. Unfortunately, procuring large training datasets often requires extensive labeling efforts in the process. Indeed, these labeling efforts can be prohibitively expensive considering the task. A prime example of this is medical imaging, where the labels are provided by trained radiologists whose services can cost well over 100 USD per hour. As such, the need for label-efficient approaches has become a hot topic for deep learning.

Active learning (AL) is one such method that addresses these labeling costs. AL methods aim to choose a cardinality-constrained subset of the unlabeled dataset whose addition to the labeled dataset after being labeled produces the largest gain in performance. Such methods do so by leveraging information contained within a working model, its training dataset, and an unlabeled dataset. Typically, such methods use this information to select a possibly diverse set of instances whose label predictions are uncertain. Examples of such methods include entropy sampling~\cite{al_survey}, which chooses unlabeled instances whose class probability distributions have maximal entropy, and \textsc{Badge}~\cite{badge}, which chooses a diverse set of uncertain instances by applying \textsc{k-means++} initialization~\cite{kmeansplus} to the space of hypothesized loss gradients of the unlabeled dataset. In general, these selected instances 
\begin{wrapfigure}[31]{r}{0.5\linewidth}
    \centering
    \includegraphics[width=\linewidth]{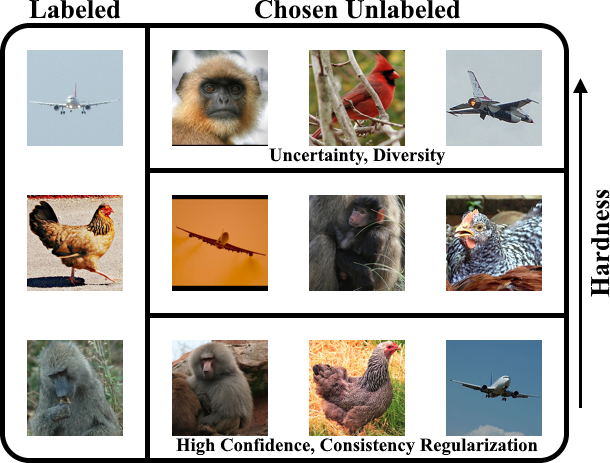}
    \caption{Depiction of the unlabeled instances selected by differing frameworks (images from STL-10~\cite{stl10}). Active Learning (AL) usually selects harder instances, often maximizing uncertainty, diversity, or both. Semi-Supervised Learning (SSL) usually selects instances whose classes are easy to predict (high confidence) or are otherwise consistent across some augmentation of the input (consistency regularization). In between these two extremes are those instances whose correct label information can impart better model performance than the auto-labeled instances of SSL while being easier to make accurate label suggestions for human labelers than pure AL.}
    \label{fig:ex_hardness}
\end{wrapfigure}
are hard to predict for the current model; thus, their inclusion in the labeled dataset tends to improve model performance after they are labeled.

While AL methods are a label-efficient way to improve model performance, most AL approaches focus on the chosen subset size as the metric of labeling cost. However, recent efforts in industry\footnote{See AWS's Amazon SageMaker Automated Labeling Service \url{https://docs.aws.amazon.com/sagemaker/latest/dg/sms-automated-labeling.html} and \url{thehive.ai}} are placing more emphasis on \emph{automating label assignment of portions of the subset selection process and enabling humans to leverage label suggestions when assigning labels}. Hence, the time that an annotator spends labeling an instance can differ depending on whether he or she can utilize a correct label suggestion or has to fix an incorrect label suggestion. As a result, this necessitates using a new labeling cost metric since the true cost of labeling is dependent not only on the number of selected instances but also on the time spent to label each, which can change depending on the correctness of the label suggestion. 


Another related paradigm is the concept of automatically labeling or pseudo-labeling unlabeled data instances, something that is followed heavily in the semi-supervised learning (SSL) literature~\cite{fixmatch}. While these auto-labeled instances come at no cost, these instances are often those that are easy to label and do not impart as much performance benefit as the harder points selected by AL do. Furthermore, there are those instances that lie within the two label hardness extremes of AL and SSL that have not been highlighted but have desirable benefits when incorporating label suggestions in the annotation process. Indeed, verifying a correctly suggested label is typically a less costly procedure than fixing an incorrectly suggested label, and the instances lying between the two hardness extremes are likely to impart more performance benefit than auto-labeled instances while retaining a significant portion of the suggested label accuracy. We summarize the relationship between all three regimes in Figure~\ref{fig:ex_hardness}. In summary, methods that factor in the benefits of all three regimes into their selection and present a meaningful computation of the labeling cost for those selected points are needed to reflect more faithful returns in model performance versus the labeling cost.

\subsection{Our Contributions}


    To confirm the claim that it is easier for a human to verify a suggested label than to correct it, we conduct a human study across four datasets to explore the values of the cost of fixing an incorrectly suggested label ($c_a$) and the cost of verifying a correctly suggested label ($c_v$). We show that $c_a:c_v$ is between $3$ to $4$, suggesting that it takes $3\times$ to $4\times$ longer to fix incorrect label suggestions (to label from scratch) than to verify correct label suggestions (see Figure~\ref{fig:c_a_c_v_values}).
    
    As it is often the case that labels can be suggested to annotators, we offer an improved baseline method for incorporating label suggestions into any AL algorithm (which we call \suggest) -- we simply suggest the model's label prediction on each selected AL instance to the annotator. By performing a simple AL experiment with \textsc{Badge}, we observe that \textsc{Badge} with \suggest\ is $1.15\times$ to $1.4\times$ more cost-efficient (label-efficient) compared to \textsc{Badge} without \suggest\ (see Figure~\ref{fig:strawman_gains}). 
    
    While the improved baseline \suggest\ is a good starting point, it does not provide significant gains for datasets that have many classes. We propose a new framework called \model, which utilizes tiered hardness by considering all three of the regimes mentioned in Figure~\ref{fig:ex_hardness} in an effort to most efficiently leverage the true cost of labeling. Our method supplements existing AL methods via the following: 1) The \suggest\ paradigm is used with existing AL methods to select hard instances with label suggestions. 2) The instances of intermediate hardness are chosen via the use of submodular mutual information (SMI)~\cite{smi}, which effectively trades off between typical measures favored by AL methods (such as diversity and uncertainty) and the relevance to the currently labeled instances while also providing a novel label suggestion scheme. Lastly, 3) our method selects the easy instances via common auto-labeling techniques to benefit from auto-labeling approaches. Notably, this framework admits most choices of AL algorithms and auto-labeling algorithms while providing a meaningful way to calculate labeling cost in accordance with recent labeling frameworks.
    
    Using our method, we show an increase of $2\%$ to $5\%$ in test accuracy over the simple AL label suggestion scheme mentioned in our second contribution across CIFAR-100~\cite{cifar}, Caltech-UCSD Birds-200-2011~\cite{caltechbirds}, and Stanford Dogs~\cite{stanforddogs} in Figure~\ref{fig:acc}. Additionally, we show labeling efficiencies of up to $2\times$ with respect to existing AL methods without label suggestions. We then study the label suggestion accuracy of each component of our method in Figure~\ref{fig:suggestion_acc}, showing that each component targets its intended regime of Figure~\ref{fig:ex_hardness}. We conclude by highlighting the importance of each component via an ablation study of our method in Figure~\ref{fig:ablation}.

\section{Related Work}
\label{sec:related_work}

Our method is positioned in a vast space of existing AL paradigms, lending part of its functionality to recent works in AL. One such work is \textsc{Badge}~\cite{badge}, which aims to select hard instances via notions of uncertainty and diversity. \textsc{Badge} selects unlabeled instances by representing each through its hypothesized loss gradients at the model's last-layer parameters, wherein the hypothesized loss is calculated using the highest-confidence class as the pseudo-label. By using \textsc{k-means++} initialization~\cite{kmeansplus} on these representations, \textsc{Badge} effectively selects a set of instances whose gradients have high magnitudes (and are thus uncertain) and are diverse. \textsc{Bait}~\cite{bait}, a recent state-of-the-art AL method, employs the use of Fisher information matrices to make informed decisions for batch selection.~\cite{bait} additionally link \textsc{Bait} as a generalization of \textsc{Badge} as its gradient embeddings correspond to rank-one approximations of \textsc{Bait}'s Fisher information matrices. Other AL methods select hard instances by directly modeling the prediction uncertainty of each unlabeled instance, such as entropy sampling, least-confidence sampling, and margin sampling~\cite{al_survey}. While the introduction of these methods are not as recent,~\cite{eval} show that the inclusion of data augmentation techniques can often bring these simple uncertainty-based techniques to match state-of-the-art AL approaches on many common image classification datasets. In our work, we utilize \textsc{Badge}~\cite{badge} and entropy sampling~\cite{al_survey} as possible choices of AL methods to be used in our framework amidst a vast space of AL paradigms.

Our chosen manner of selecting the instances of intermediate hardness also utilizes submodular information measures~\cite{smi}, which have recently been used to perform a wide variety of selection tasks in AL. In particular, submodular mutual information (SMI) has been used to perform targeted subset selection via a simple cardinality-constrained optimization method detailed in~\cite{nemhauser}. To capture a variety of desirable aspects of the selected subset of targeted instances,~\cite{prism} study the use of a wide variety of SMI instantiations, offering insights to the utility of each in the selection process.~\cite{similar} utilize many of the SMI instantiations offered in~\cite{prism} and the hypothesized gradient functionality of \textsc{Badge}~\cite{badge} to perform targeted AL in the presence of rare classes. Lastly,~\cite{talisman} apply the use of SMI for selecting rare-slice images to label in the context of object detection. In our work, we utilize SMI~\cite{smi} to query for instances of intermediate hardness that are related to a labeled instance subset of uniform class as a means of procuring suggested labels (detailed in Section~\ref{sec:method}).

\section{The Benefit of Label Suggestions}
\label{sec:suggested_labels}

To further motivate our framework, we more closely examine the cost of labeling in recent pipelines, where label suggestions are provided to the human annotators. To better capture the cost of labeling in pipelines with label suggestions, we model the labeling cost as $c_v \times n_{correct} + c_a \times (n - n_{correct})$, where $c_v$ is the cost to verify a correct suggested label and $c_a$ is the cost to assign a corrected label when an incorrect suggestion is given. Recent AL works have framed the cost of labeling in terms of the number of selected unlabeled instances; however, the cost of the final assignment of an instance's label fluctuates depending on the suggested label. Instances with correct label suggestions require less time in assigning final labels ($c_v$) versus those with incorrect label suggestions ($c_a$) due to the extra correction effort. Hence, to get the best returns in model performance versus the labeling cost, AL methods must select points that impart the most information to the model while also providing accurate suggested labels.

To better understand the impact of suggested label verification on the model performance gain of these pipelines, further study about the values of $c_a$ and $c_v$ is needed. Indeed, the value of each depends on an annotator's proficiency in using the labeling pipeline, his or her knowledge of the target domain, and inherent properties of the data itself. To this end, we conduct a labeling experiment to ascertain the ranges of values that $c_a$ and $c_v$ can take for different datasets. In our experiment, voluntary human subjects utilize a labeling tool that iterates over images in the CIFAR-10~\cite{cifar}, STL-10~\cite{stl10}, SVHN~\cite{svhn}, and UC Merced Land Use~\cite{ucmerced} datasets. The subject is asked to label 100 images of each dataset (105 images for UC Merced Land Use) before moving on to the next dataset. During the labeling process, the tool offers label 
\begin{wrapfigure}[25]{r}{0.5\linewidth}
    \centering
    \includegraphics[width=\linewidth]{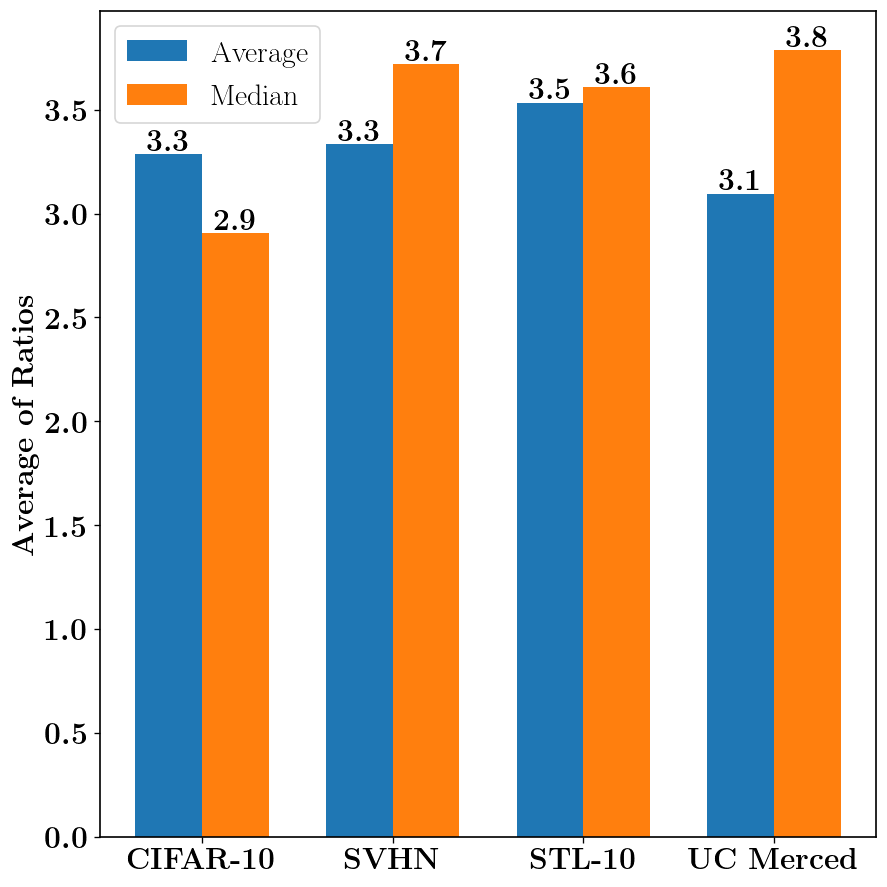}
    \caption{Average of each subject's average/median ratios for $c_a$ and $c_v$ across each of the four datasets examined in our labeling experiment. We see that most datasets admit a $3 - 4 \times$ increase in $c_a$ vs $c_v$.}
    \label{fig:c_a_c_v_values}
\end{wrapfigure}
\begin{wrapfigure}[18]{r}{0.5\textwidth}
    \centering
    \includegraphics[width=\linewidth]{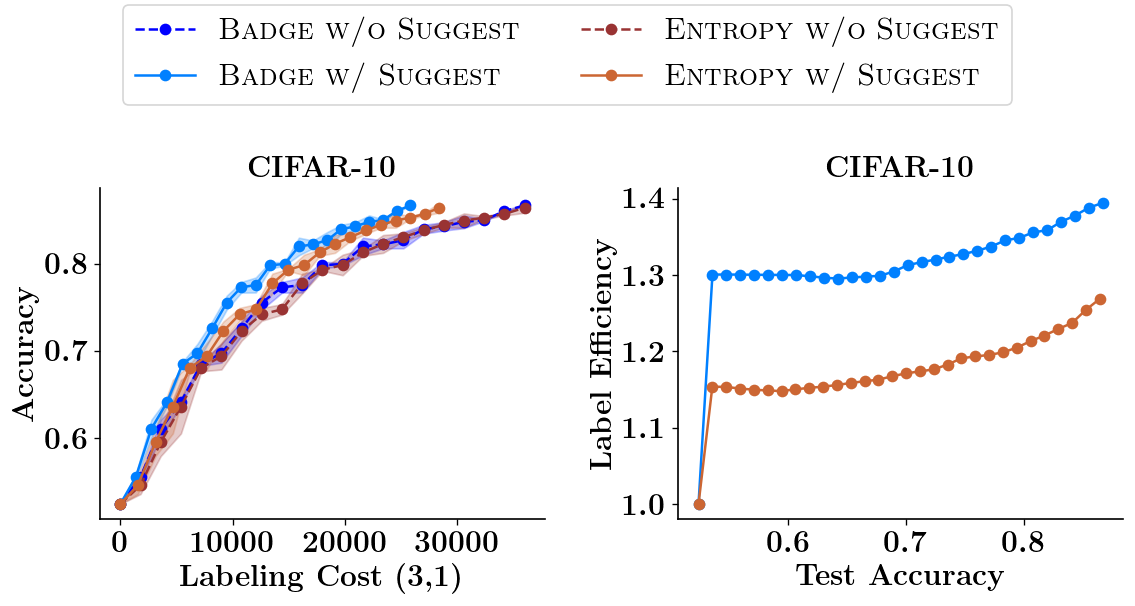}
    \caption{Improvement in test accuracy and labeling efficiency on CIFAR-10 by introducing label suggestions. Here, we see labeling efficiencies between $1.15\times$ and $1.4\times$ versus AL baselines without label suggestions when $c_a=3,c_v=1$.}
    \label{fig:strawman_gains}
\end{wrapfigure}
suggestions for each image, where the suggested label has a $50\%$ chance of being the correct ground truth label. The time taken to label each image is recorded, and images that are incorrectly labeled by the subject are discarded. Using the correctly labeled images, the average and median times to verify correctly suggested image labels are calculated, which we pose as the subject's possible values of $c_v$. Likewise, the average and median times to fix incorrectly suggested image labels are calculated, which we pose as the subject's possible values of $c_a$. More details about the experiment and the labeling tool used can be found in Appendix~\ref{app:labeling}.

\subsection{An Improved Baseline: \suggest}

In order to effectively enable suggestions in existing active learning approaches, we propose \suggest, a simple baseline that essentially uses the current model's label prediction as the suggested label for each of the selected AL instances. To visualize the effect that the \suggest\ paradigm has on existing AL methods, we run an experiment on CIFAR-10~\cite{cifar} to show the better gains in model performance and labeling efficiency that AL with label suggestions has with respect to AL without label suggestions. Here, we define labeling efficiency as the ratio between the labeling costs needed for two AL methods to reach a desired test accuracy as is done in~\cite{eval}. 

Using appropriate choices of $c_a$ and $c_v$ obtained from our labeling experiment, we can now give more faithful estimates of the true cost of labeling in pipelines with automatic label suggestions. Based on the results given in Figure~\ref{fig:strawman_gains}, we can conclude that there is a benefit in applying automated label suggestions for leveraging more out of the human labeling effort. While these AL methods improve in their labeling efficiency, however, they do not consider leveraging \emph{more} of the human labeling effort via label suggestions by targeting instances that admit easier label suggestions while retaining some of the hardness properties that they use in their selection. We expand upon this idea to present our \model\ framework in Section~\ref{sec:method}, comparing against the \suggest\ baseline in Section~\ref{sec:experiments}.

\section{Our Proposed Framework: \model}
\label{sec:method}

Here, we present \model, a new Interactive Learning framework for leveraging the most out of the human labeling effort via automatic label suggestions. In particular, our framework presents three selection methodologies for selecting instances that span across the degrees of label suggestion hardness as detailed in Figure~\ref{fig:ex_hardness}. Notably, our framework utilizes the trade-off between query relevance and uncertainty when selecting instances to label. Instances that are relevant to the labeled set (e.g., are similar to the labeled set) naturally admit confident label predictions but tend to not impart new knowledge for improving model performance. Instances that are \emph{not} relevant to the labeled set are naturally harder to classify correctly (as the prediction is more uncertain) but tend to impart more knowledge for improving model performance. By varying the degree at which query relevance and prediction uncertainty are used, a selection method can effectively target points within one of the hardness tiers mentioned in Figure~\ref{fig:ex_hardness}. By choosing instances from all hardness tiers, our method enjoys the benefits that each tier provides in the context of AL with automatic label suggestions, which we detail in this section. We summarize our framework in Algorithm~\ref{alg:framework}.

\begin{algorithm}
    \caption{\model}
    \label{alg:framework}
    \begin{algorithmic}[1]
        \STATE {\bfseries Input:} Labeled set $\mathcal{L}$, Unlabeled set $\mathcal{U}$, Model $M$, Model parameters $\theta$, Hard budget $b_1$, Intermediate budget $b_2$, Easy budget $b_3$
        \REPEAT
            \STATE $X_{hard}, \hat{\mathbf{y}}_{hard} \leftarrow \text{AL\_Select}(\mathcal{L}, \mathcal{U}, M, \mathcal{\theta}, b_1)$
            \STATE $X_{int}, \hat{\mathbf{y}}_{int} \leftarrow \text{SMI\_Suggest}(\mathcal{L}, \mathcal{U}, M, \mathcal{\theta}, b_2)$
            \STATE $X_{easy}, \mathbf{y}_{easy} \leftarrow \text{Auto\_Label}(\mathcal{L}, \mathcal{U}, M, \mathcal{\theta}, b_3)$
            \STATE $\mathbf{y}_{hard}\leftarrow \text{Label}(X_{hard}, \hat{\mathbf{y}}_{hard})$
            \STATE $\mathbf{y}_{int}\leftarrow \text{Label}(X_{int}, \hat{\mathbf{y}}_{int})$
            \STATE $X \leftarrow X_{hard} \cup X_{int} \cup X_{easy}$
            \STATE $\mathbf{y} \leftarrow \mathbf{y}_{hard} \cup \mathbf{y}_{int} \cup \mathbf{y}_{easy}$
            \STATE $\mathcal{U} \leftarrow \mathcal{U} - X$
            \STATE $\mathcal{L} \leftarrow \mathcal{L} \cup \left[\bigcup_{i=1}^{|X|} (X_i,\mathbf{y}_i)\right]$
            \STATE $\theta \leftarrow \text{Train}(\mathcal{L},M,\theta$)
        \UNTIL converged
    \end{algorithmic}
\end{algorithm}

\paragraph{Selecting Hard Instances (AL\_Select):} While choosing instances that are relevant to the labeled set admits more accurate label suggestions, strategies that over-focus on maintaining highly confident label suggestions tend to select instances that are not as informative for model training as other instances in the unlabeled set. For this reason, using existing AL methods for selecting hard instances is favorable for maintaining consistent improvement in model performance. As previously discussed, however, selecting too many hard instances \emph{en masse} tends to neglect the benefit of reduced labeling cost granted by correct label verification. As such, using existing AL methods to select hard instances can be done in moderation to glean important knowledge from the unlabeled set while being cognizant of the labeling cost. In our experiments, we examine the use of \textsc{Badge}~\cite{badge} and entropy sampling~\cite{al_survey} as the AL component of our framework.

\paragraph{Selecting Intermediate Instances (SMI\_Suggest):} As previously mentioned, selecting too many hard instances misses out on the benefit of correct label verification. On the other hand, selecting too many easy instances misses out on the knowledge gained from the hard instances. To gain benefits from both aspects, instances that lie between these two hardness extremes are desired. To target these instances, methods that factor in relevance to the labeled set and notions of uncertainty can be used. To that end, we utilize the recently proposed submodular mutual information~\cite{smi} to select instances that fall between both hardness extremes and to provide their suggested labels. Submodular mutual information (SMI), denoted $I_f(A;Q)$, measures the overlap in information between two sets of instances $A,Q$ with respect to a submodular information function $f$. By selecting $f$ to model diversity information (to get instances that are not over-focused on relevance), one can capture both query relevance and diversity/uncertainty information through $I_f(A;Q)$. To select instances from the unlabeled set and to provide their suggested labels, the following cardinality-constrained optimization problem is solved on a per-class basis using the simple greedy algorithm proposed by~\cite{nemhauser}:
\begin{align}
    \max_{A_c\subseteq \mathcal{U},|A_c|\leq k} I_f(A_c;\mathcal{L}_c)
\end{align}
\noindent where $\mathcal{L}_c$ denotes the set of labeled instances belonging to class $c$. Hence, each selected subset $A_c$ is a set of unlabeled instances that have query relevance to $\mathcal{L}_c$; accordingly, we assign a suggested label $c$ for each point in $A_c$. Each $A_c$ (along with each suggested label) is coalesced to form the set of selected instances of intermediate hardness. For our experiments, we utilize \textsc{LogDetMI}~\cite{smi,prism,similar} to capture query relevance and diversity/uncertainty information. More details can be found in Appendix~\ref{app:smi}.

\paragraph{Selecting Easy Instances (Auto\_Label): } The last component of our framework addresses those instances that are highly relevant to the labeled set. In many cases, there are cases where the model or some other mechanism can make highly confident label predictions for some of the unlabeled instances. The high-confidence of these instances tends to originate from high relevance to the labeled set, so these instances tend to not impart as much knowledge for improving model performance. However, as these instances admit high-confidence predictions, one can choose to automatically assign the suggested label as the ground-truth label instead of presenting them to human labelers, which may present a waste of labeling effort. Hence, these instances can be added to the labeled set at no additional cost, imparting some additional knowledge for improving the model. Such selection is not without risk, however, as there is a risk of auto-assigning noisy labels; hence, the budget of this component of our framework can be attenuated according to the accuracy of the chosen auto-labeling method. In our experiments, we select the easy instances by choosing those instances in the unlabeled set with the highest-confidence predictions.

\section{Experiments}
\label{sec:experiments}

In this section, we conduct experiments to answer the following questions: \textbf{1)} Utilizing our method, what is the improvement in accuracy over AL versus the labeling cost discussed in Section~\ref{sec:suggested_labels}? \textbf{2)} How hard is it to predict the correct label for the instances selected by each component; \emph{e.g.}, is it true that the label hardness presented in Figure~\ref{fig:ex_hardness} is correct? Lastly, \textbf{3)} Are the different components of our method necessary? We analyze our method across three datasets -- CIFAR-100~\cite{cifar}, Caltech-UCSD Birds-200-2011~\cite{caltechbirds}, and Stanford Dogs~\cite{stanforddogs} -- to answer each.

\subsection{Experimental Setting}

Here, we enumerate various aspects of the experimental setting used in our experiments. Additional reproducibility information is given in Appendix~\ref{app:reproducibility}.

\paragraph{Compute and Datasets: } To run our experiments, we utilize a machine with 96 GB of RAM, an NVIDIA RTX A6000 GPU with 48 GB of VRAM, and an Intel Xeon Silver 4208 CPU. To effectively utilize all these components, we implement our code in PyTorch~\cite{pytorch}. Using this architecture, we conduct experiments across three datasets: CIFAR-100~\cite{cifar}, Caltech-UCSD Birds-200-2011~\cite{caltechbirds}, and Stanford Dogs~\cite{stanforddogs}. CIFAR-100 is an image classification dataset of 60k images spread across 100 classes. Caltech-UCSD Birds-200-2011 is a fine-grained image classification dataset of approximately 12k bird images, each of which is assigned one of 200 possible species. Lastly, Stanford Dogs is another fine-grained image classification dataset of approximately 21k dog images, each of which is assigned one of 120 possible breeds.

\paragraph{Model Architecture: } 
For each dataset, we utilize the ResNet-18~\cite{resnet} architecture for image recognition. We reset the model parameters after each round of labeling following the observation that resetting parameters leads to better generalization performance as detailed in~\cite{eval}. However, both the Caltech-UCSD Birds-200-2011 and Stanford Dogs datasets do not admit enough data to fully train all parameters of the ResNet-18 model without severe underfitting. To rectify this, we opt to use pre-trained feature extractor weights obtained from PyTorch's collection of pre-trained models for these two datasets. These weights are kept frozen throughout each experiment on these two datasets; only the fully connected layer is reset and trained after each selection.

\paragraph{Training: } To train each model between labeling rounds, we utilize PyTorch's SGD optimizer with momentum ($\mu=0.9$) and weight decay ($\lambda=5\times 10^{-4}$). Additionally, we use a cosine-annealing learning rate with an initial setting of $\gamma = 0.01$. $T_{max}$ is set to the maximum number of epochs that we allow in training each model, which we set to $300$ for CIFAR-100 and $500$ for Caltech-UCSD Birds-200-2011 and Stanford Dogs. Additionally, we stop training the model parameters once training accuracy has reached $99\%$ on CIFAR-100 and $95\%$ on the other datasets. We employ data augmentations during training to leverage better model generalization as is discussed in~\cite{eval}. On CIFAR-100, we utilize random crops, random horizontal flips ($p=0.5$), and normalization for training; we utilize only normalization during testing. For the other datasets, we resize each image to a $256\times 256$ dimension before applying random cropping (to $224\times 224$ images), random horizontal flips ($p=0.5$), and normalization for training; we resize each image to a $256\times 256$ dimension, apply center cropping (to $224\times 224$ images), and normalization for testing.

\begin{figure*}[t]
    \centering
    \includegraphics[width=\linewidth]{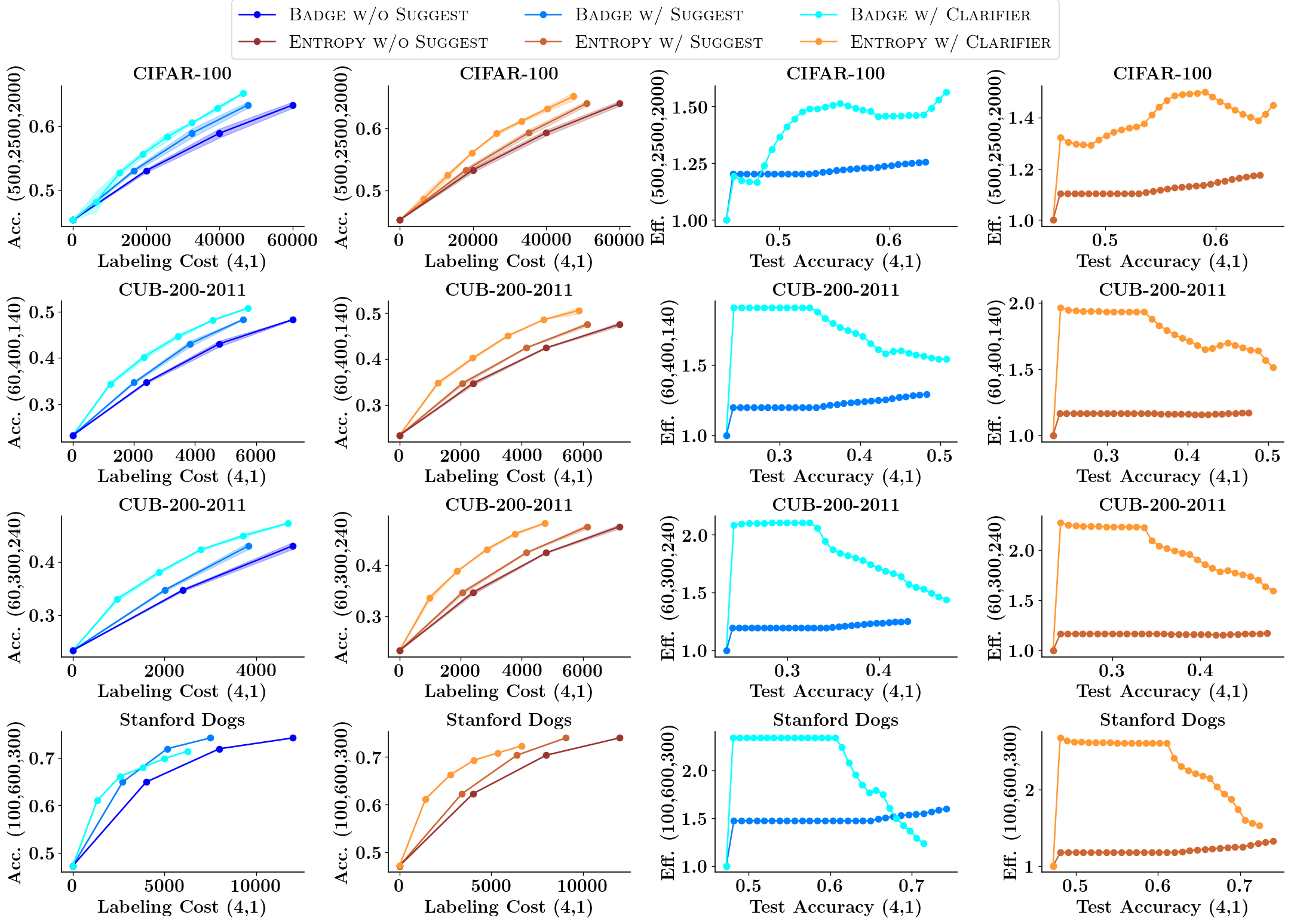}
    \caption{Accuracies on all four datasets. Using \model\ shows improvement over existing AL methods and the \suggest\ baseline discussed in Section~\ref{sec:suggested_labels}. The labeling cost is calculated based on $c_a$ and $c_v$ values that are observed in Section~\ref{sec:suggested_labels}.}
    \label{fig:acc}
\end{figure*}

\paragraph{Selection Partitioning: } While some selection methods are effective at choosing hard (\textsc{Badge}) or otherwise targeted (\textsc{SMI}) instances, the size of the unlabeled set often imposes prohibitive memory requirements on some systems. Following the strategy used in~\cite{similar}, we randomly partition the unlabeled set into equal-sized chunks and perform selections on each, equally partitioning the selection budget among each chunk. In doing so, these methods that would otherwise not scale effectively to the large unlabeled set can be utilized on reasonably sized portions of the unlabeled set. On CIFAR-100, we partition the unlabeled dataset into 5 partitions when selecting intermediate instances with SMI. Additionally, we partition the dataset into 5 partitions when using AL to select instances in CIFAR-100. Otherwise, we do not perform any other partitioning when selecting instances to label.

\paragraph{Seed Size and Budgets: } To begin each experiment, we choose to start with labeled seed sets of size 10k, 600, and 500 for CIFAR-100, Caltech-UCSD Birds-200-2011, and Stanford Dogs, respectively. We then choose values of $b_1,b_2$, and $b_3$ as described in Algorithm~\ref{alg:framework} for each dataset, which are listed in order in parentheses for each plot in Figure~\ref{fig:acc}. For the compared AL baseline in each plot, we utilize a budget equal to the sum of $b_1,b_2$, and $b_3$ so that the same number of instances are added per labeling round.

\newpage
\subsection{Accuracy Improvement}

Here, we present the results of each experiment. To produce error bars, we run each compared method 3 times across CIFAR-100. We run each compared method 2 times for Caltech-UCSD Birds-200-2011 and 1 time for Stanford Dogs. We show improvement in test accuracy versus labeling cost in Figure~\ref{fig:acc}, along with improvement in labeling efficiency (see Section~\ref{sec:suggested_labels}). Utilizing the labeling experiment results in Section~\ref{sec:suggested_labels}, we assume $c_a,c_v$ values as stated in the parentheses in Figure~\ref{fig:acc} to calculate the labeling cost. \model\ enjoys accuracy improvements of up to $3\%$ on CIFAR-100 and $5\%$ on Caltech-UCSD Birds-200-2011 and Stanford Dogs. Accordingly, the labeling efficiencies for these datasets range from $1.0\times$ to $2.5\times$ when comparing \model\ to existing AL approaches without label suggestions, demonstrating more efficient use of the human labeling effort (even compared to the same AL approaches with label suggestions given by \suggest). We conclude that our method returns more improvement in model performance versus the labeling cost, especially for datasets with large class counts.

\begin{figure*}[t]
    \centering
    \includegraphics[width=0.95\linewidth]{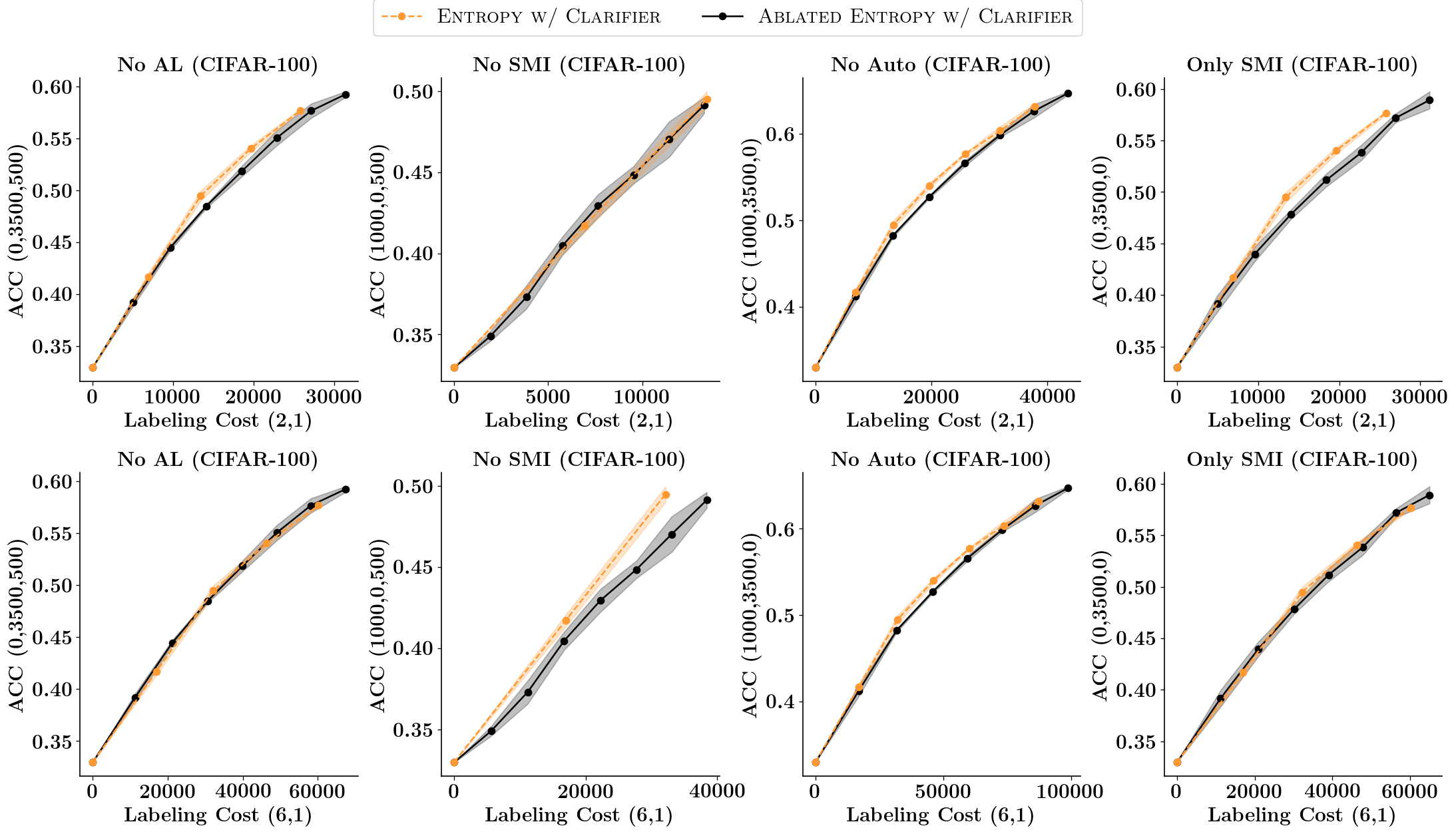}
    \caption{Ablation studies across CIFAR-100 using an initial labeled seed set of 5k instances. Removing a component from \model\ results in poorer test accuracy improvement versus the labeling cost. The AL component is more useful with lower $c_a:c_v$; the SMI component is more useful with higher $c_a:c_v$; the auto-labeling component helps improve accuracy at no additional cost.}
    \label{fig:ablation}
\end{figure*}

\subsection{Label Suggestion Hardness} 

\begin{wrapfigure}[19]{r}{0.5\linewidth}
    \centering
    \includegraphics[width=\linewidth]{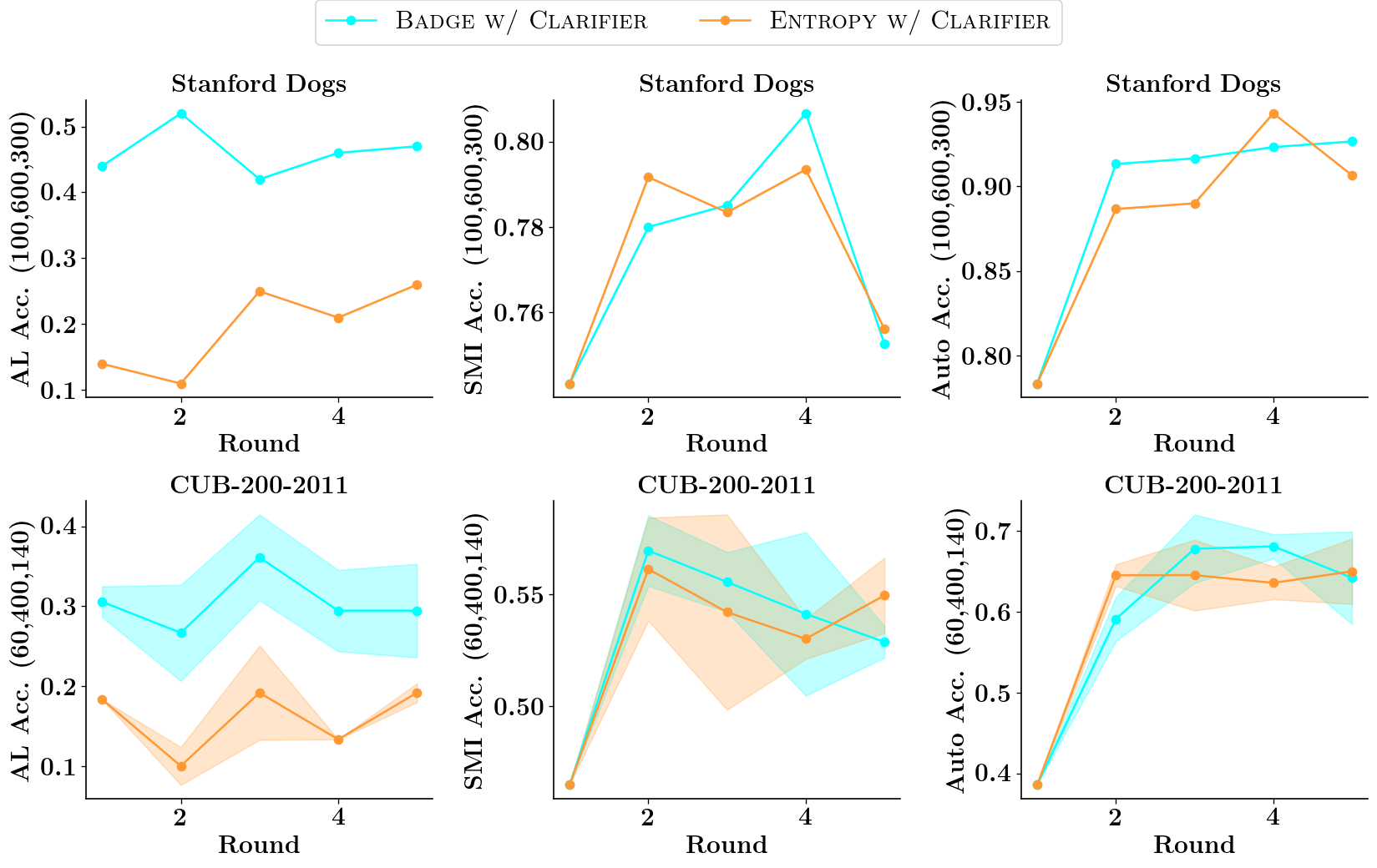}
    \caption{Suggestion accuracies on Stanford Dogs and Caltech-UCSD Birds-200-2011. Across each experiment, the AL methods selects the hardest instances, followed by SMI. Auto-labeling selects the easiest examples to label.}
    \label{fig:suggestion_acc}
\end{wrapfigure}

Next, we study the hardness of the suggested label for each of the domains presented in Figure~\ref{fig:ex_hardness}. To do so, we present the label suggestion accuracy of each component during each labeling round of Caltech-UCSD Birds-200-2011 and Stanford Dogs in Figure~\ref{fig:suggestion_acc}. Interestingly, we find that \textsc{Badge} enjoys better suggestion accuracy over entropy sampling, which most likely occurs since entropy sampling directly selects those instances whose class probabilities are most uncertain. As shown in each column, the label suggestion accuracy for instances selected by AL methods is consistently less than that for instances selected by \textsc{LogDetMI}. Likewise, the label suggestion accuracy for instances selected by \textsc{LogDetMI} is consistently less than that for instances 
selected by the auto-labeler. This verifies our claim that each component selects instances from its targeted domain as portrayed in Figure~\ref{fig:ex_hardness}.

\subsection{Ablation} 

Lastly, we study the effect that each component has on the returns in model performance versus the labeling cost. We present ablated versions of a CIFAR-100 experiment using the \textsc{Entropy+\allowbreak Logdetmi+Highest-confidence} configuration of \model\ in Figure~\ref{fig:ablation}. In each, we examine the effect of setting the selection budget of each component to 0. Note that we already present pure AL methods in Figure~\ref{fig:acc}. Furthermore, we do not present the configuration where only auto-labeled instances are assigned. Such configurations have been more heavily studied in SSL, and since our auto-labeling component makes hard assignments on easy instances, the test accuracy is not likely to improve significantly. Lastly, we present these ablated versions using multiple configurations for $c_a$ and $c_v$ to further analyze the contribution of each component in different cost settings.

Across all configurations, we see that the performance of the ablated version either matches that of the original \textsc{Entropy+Logdetmi\allowbreak +Highest-confidence} \model\ configuration or is below it. Notably, Figure~\ref{fig:ablation} suggests that the $c_a:c_v$ ratio directly influences the model performance gain versus the labeling cost of each component of \model. Namely, we observe that the AL component (which selects the hard instances) is more beneficial in leveraging the labeling cost when the $c_a:c_v$ ratio is small since its removal results in a noticeable performance degradation. When the $c_a:c_v$ ratio is increased, this performance degradation is nullified, suggesting that the AL component becomes less beneficial as the $c_a:c_v$ increases. We observe the opposite behavior with the SMI component (which selects the intermediate instances). Indeed, there is a noticeable performance degradation when removing SMI in the higher $c_a:c_v$ setting presented in Figure~\ref{fig:ablation}, but this performance degradation is nullified in the smaller $c_a:c_v$ setting. Lastly, we observe that there is a slight performance degradation when removing the auto-labeling component, which is unaffected by the $c_a:c_v$ ratio. We conclude that each component is necessary to fully leverage the human labeling effort to maximize the gain in model performance across multiple cost scenarios.

\section{Discussion}
\label{sec:discussion}

In this work, we more closely examine the need for Interactive Learning frameworks that match the trend of increased use of auto-labeling. Accordingly, we frame this increased use of AL methods with auto-labeling in the context of label prediction hardness. Through a human labeling experiment, we show that verifying correctly suggested labels is less costly than fixing incorrectly suggested labels, highlighting a potential source of improvement in gaining model performance versus the true human labeling cost. By combining this result with the observation of label prediction hardness, we formulate \model, a new Interactive Learning framework that targets instances from each hardness tier to leverage better model performance gains out of the labeling effort. By our experiments, we show that this framework achieves better gains in accuracy over existing AL methods and reinforces our view of label prediction hardness. 

As future work, more adaptive choices of each component's budget can be studied to leverage even more control of the labeling effort as the choice of the budget parameters are not studied heavily, which limits the direct application of \model. We also stress that the use of \model\ is more applicable with large class counts and is highly dependent on the human component: those annotators with low $c_a : c_v$ ratios will not see marked benefit in using \model. As we view \model\ as a foundational work in this setting, we do not anticipate any clear negative societal impacts.

\bibliographystyle{unsrt}  
\bibliography{main}  

\clearpage
\newpage
\appendix

\begin{center}
\part{Supplementary Material for Beyond Active Learning: Leveraging the Full Potential of Human Interaction via Auto-Labeling, Human Correction, and Human Verification} 
\end{center}
\parttoc 

\section{Reproducibility}
\label{app:reproducibility}

While we mention the experimental setting in Section~\ref{sec:experiments}, we discuss a couple other aspects of reproducibility in this section. We provide an experiment script\footnote{\url{https://github.com/nab170130/auto_label_mp}} that executes a \model\ configuration given a set number of command-line arguments:

\begin{itemize}
    \item \textbf{al\_strategy}: One of \textit{badge, entropy}
    \item \textbf{auto\_assign\_strategy}: \textit{highest\_confidence}
    \item \textbf{b1}: AL selection budget
    \item \textbf{b2}: SMI selection budget
    \item \textbf{b3}: Auto-assign budget
    \item \textbf{dataset}: One of \textit{cifar10, cifar100, birds, dogs}
    \item \textbf{device}: CUDA device ID
    \item \textbf{human\_correct\_strategy}: \textit{logdetmi}
    \item \textbf{num\_partitions\_human}: Number of unlabeled set partitions (see Section~\ref{sec:experiments})
    \item \textbf{rounds}: Number of selection rounds
    \item \textbf{runs}: Number of repeated experiment runs to execute
    \item \textbf{seed\_size}: Size of the initial labeled seed set
    \item \textbf{thread\_count}: Number of threads in SMI selection
\end{itemize}

\noindent The script utilizes the DISTIL toolkit presented in~\cite{eval} to produce result JSON files that contain information about each selection round. We utilize a separate Jupyter notebook to analyze and plot our results, which we provide with the experiment script. Lastly, we detail the licenses of each repository and dataset used in this work:

\begin{itemize}
    \item Caltech-UCSD Birds-200-2011~\cite{caltechbirds}: Non-commercial Research
    \item CIFAR-10(0)~\cite{cifar}: MIT License
    \item DISTIL~\cite{eval}: MIT License
    \item labelImg~\cite{labelimg}: MIT License
    \item PyTorch~\cite{pytorch}: Modified BSD
    \item Stanford Dogs~\cite{stanforddogs}: Non-commercial Research
    \item STL-10~\cite{stl10}: Non-commercial Research
    \item SVHN~\cite{svhn}: CC0 1.0 Public Domain
    \item UC Merced Land Use~\cite{ucmerced}: CC0: Public Domain
\end{itemize}

\section{Additional Labeling Experiment Details}
\label{app:labeling}

\begin{figure*}[t]
    \centering
    \includegraphics[width=\linewidth]{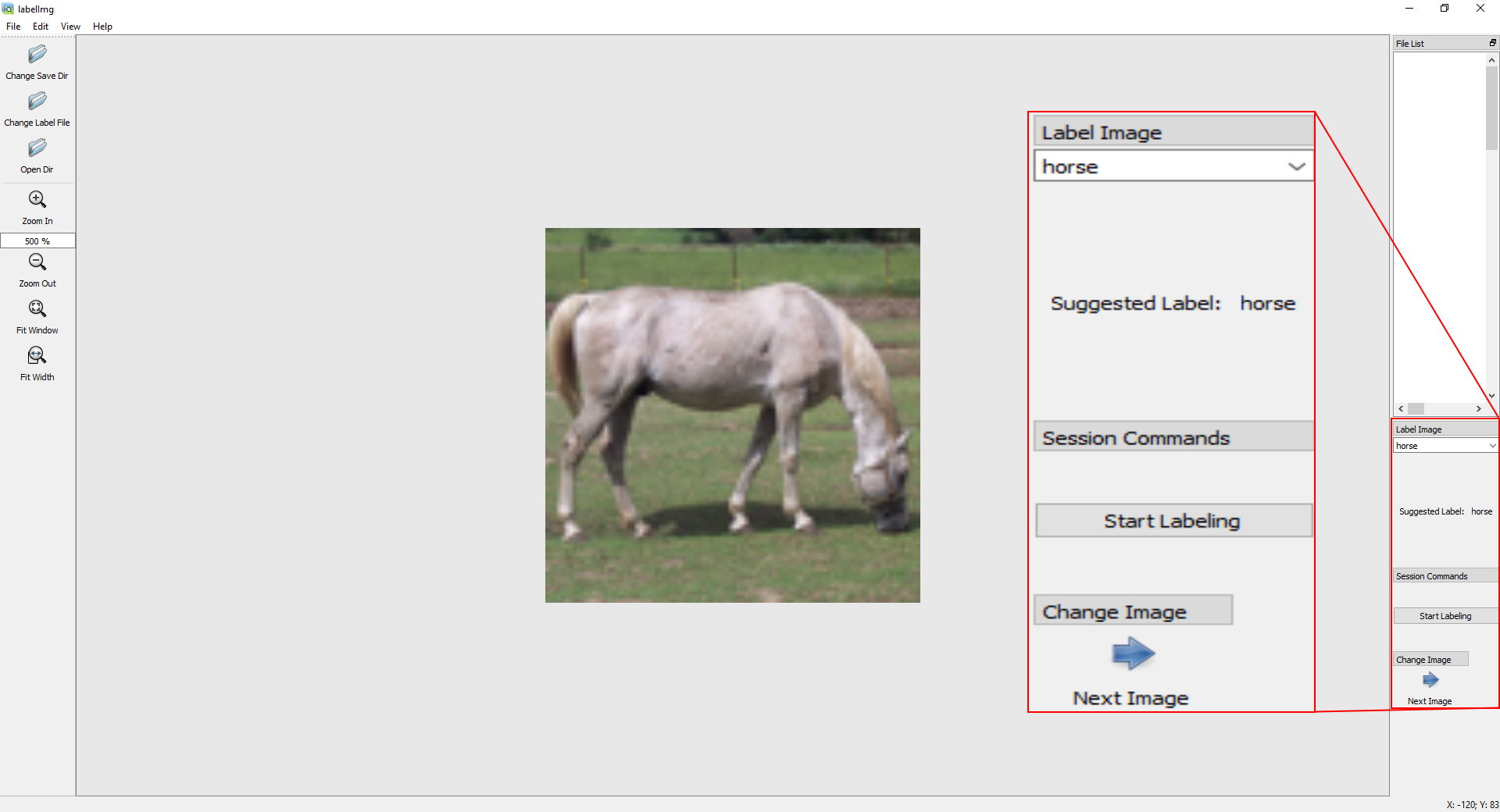}
    \caption{The tool interface used in our labeling experiment. The subject presses the "Start Labeling" button to initiate the timing experiment. The tool provides a suggested label that is correct $50\%$ of the time. The subject fixes the suggested label by either selecting the correct label from the drop-down menu or by typing the label in the drop-down menu's field. The subject clicks the "Next Image" button once he or she has verified the correctness of the suggested label or has fixed an incorrect suggested label.}
    \label{fig:tool}
\end{figure*}

Here, we present more details on the labeling tool used in our labeling experiment. 

\subsection{Labeling Tool}

To compute estimates of the cost to fix incorrectly suggested labels ($c_a$) and the cost to verify correctly suggested labels ($c_v$) for each of the datasets discussed in Section~\ref{sec:suggested_labels}, we utilize a labeling tool based on~\cite{labelimg} that records the time spent to assign a final label for a series of images with potentially incorrect suggestions. A snapshot of the tool is given in Figure~\ref{fig:tool}. Once a dataset has been chosen, the tool sequentially presents a random sample of images in that dataset. Timing information starts to be collected when the user presses the "Start Labeling" button. From that point, the subject assigns the final label for the image by pressing the "Next Image" button, which immediately presents the next image. The subject is provided a suggested label for each image to assist in the labeling effort. If the suggestion is wrong (which occurs $50\%$ of the time as mentioned in Section~\ref{sec:suggested_labels}), then the subject can correct the suggestion by selecting the correct label from the drop-down menu or by typing the label in the drop-down menu's field. Once the subject has finished labeling the image sample for a dataset, the tool saves the timing information for each image along with the final label given by the subject. 

\begin{figure*}
    \centering
    \includegraphics[width= \linewidth]{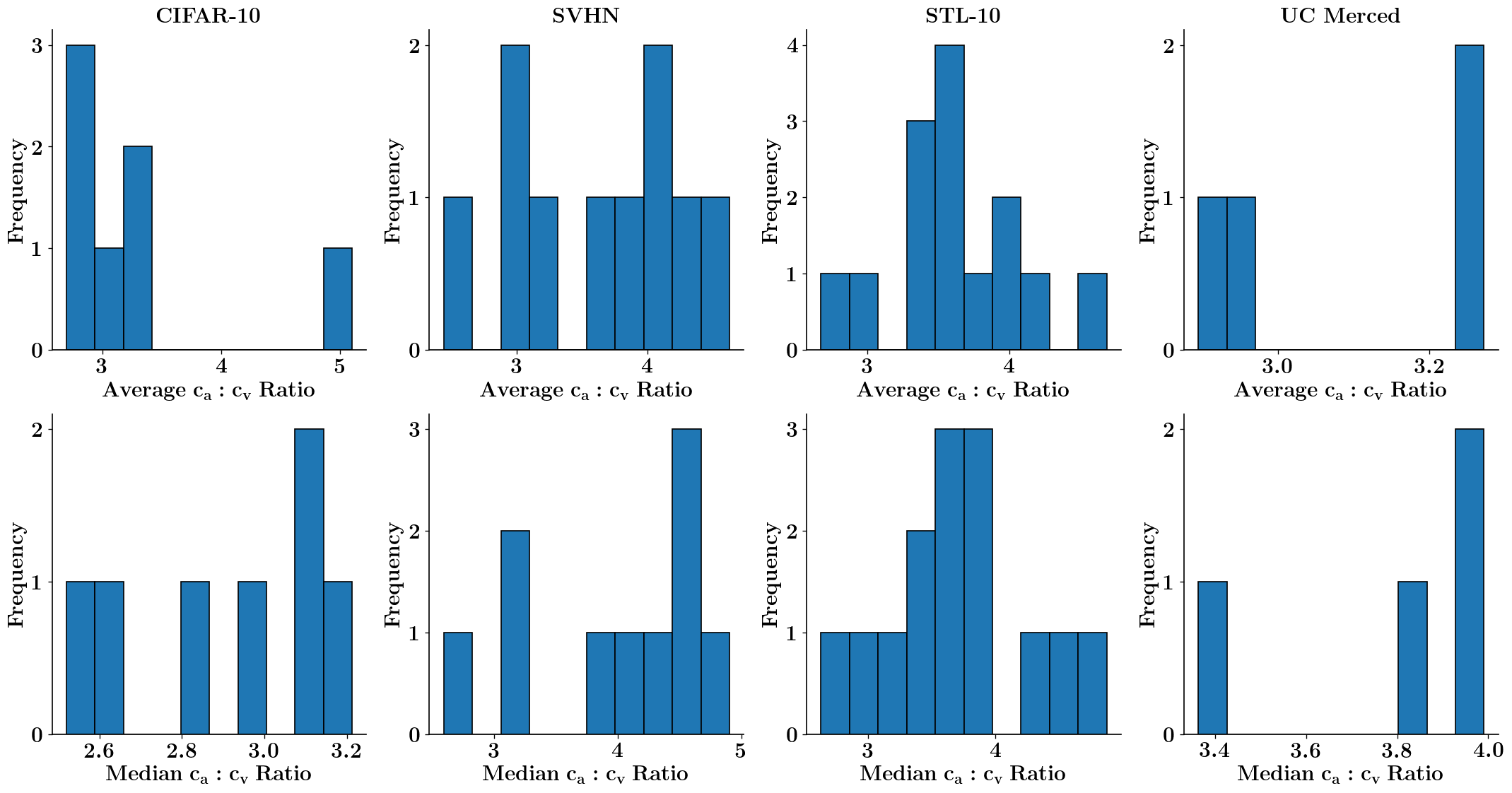}
    \caption{Histograms of average / median ratios across each dataset. As shown, the $c_a:c_v$ ratio for most subjects falls between $3$ and $4$.}
    \label{fig:ratio_histogram}
\end{figure*}

When computing the average and median values for $c_a$ and $c_v$ for each subject's performance on a dataset, only the images whose final label matches the ground truth label of the dataset are considered. These average and median values are also saved and are used to produce average and median ratios between $c_a$ and $c_v$ for a subject's performance on a dataset. In Section~\ref{sec:suggested_labels}, the average of the average ratios and the average of the median ratios are presented in Figure~\ref{fig:c_a_c_v_values}. We present the distribution of average and median $c_a:c_v$ ratios for each dataset in Figure~\ref{fig:ratio_histogram}.

\section{SMI-Based Selection}
\label{app:smi}

In this section, we highlight the optimization background and the methodology used in selecting the intermediate-hardness instances mentioned in Figure~\ref{fig:ex_hardness} and in Section~\ref{sec:method}.

\subsection{Submodular Mutual Information} 

To target the intermediate-hardness instances, we apply the recently proposed submodular mutual information~\cite{smi}, which we introduce and motivate here. A set function $f:2^\Omega \rightarrow \mathbb{R}$ is submodular if $f(A \cup \{x\}) - f(A) \geq f(B \cup \{x\}) - f(B)$ for any $A\subseteq \Omega,B\subseteq \Omega,x \in \Omega$ satisfying $A \subseteq B, x \notin B$. Furthermore, $f$ is monotone if $f(A) \leq f(B)$ for any $A\subseteq B\subseteq \Omega$. If $f$ is both monotone and submodular, then the function can be maximized under a cardinality constraint on its domain with a $(1-\frac{1}{e})$ approximation guarantee using a simple greedy algorithm~\cite{nemhauser}. In many instances, real-valued set functions are used to assign scores to subsets that correspond to ratings of set coverage, set diversity, and other metrics of interest; furthermore, many of these set functions are monotone submodular, giving a very sound method of selecting cardinality-constrained subsets from large amounts of data that appeal to some desired measure. 

Submodular functions have also been used to model information-theoretic quantities.~\cite{smi} define the submodular mutual information between sets $A,Q\subseteq \Omega$ under $f$ as $I_f(A;Q) = f(A) + f(Q) - f(A \cup Q)$. In addition to being monotone,~\cite{smi} show that $I_f(A;Q)$ is also submodular in $A$ or $Q$ when fixing the other argument under further conditions on $f$. Hence, $I_f(A;Q)$ can be fixed in $Q$ and maximized by the aforementioned greedy algorithm~\cite{nemhauser} for sufficient $f$, giving cardinality-constrained subsets that are similar to $Q$ under $f$ that also appeal to the measure modeled by $f$. As such, $Q$ can be treated as a query set that represents a target for the selected subset. This strategy has been used in prior work to perform targeted AL; namely,~\cite{similar} use various submodular mutual information instantiations as described in~\cite{prism} to mine rare-class, non-redundant, and in-distribution instances for image classification. \cite{talisman} use these functions for mining rare objects and slices in object detection problems.

%
%


\subsection{Choice of SMI Function} 

As detailed in~\cite{similar} and~\cite{prism}, many different instantiations of submodular mutual information can be chosen. In this work, we opt to instantiate $I_f$ as the Log-Determinant Mutual Information (\textsc{LogDetMI}) function used in~\cite{similar} and~\cite{prism}:

\begin{align}
    I_f(A;Q) = \log \det \mathcal{S}_A - \log \det \left(\mathcal{S}_A - \mathcal{S}_{A,Q}S^{-1}_QS^T_{A,Q}\right)
\end{align}

\noindent where $\mathcal{S}_A$ is a matrix of similarity scores between the instances in $A$ and $\mathcal{S}_{A,Q}$ is a matrix of similarity scores between the instances in $A$ to the instances in $Q$. To compute each $(S_{A,Q})_{ij}$, we take the cosine similarity between their vector representations $v_i,v_j$. We formulate $v_i,v_j$ in the same manner done by~\cite{badge} and~\cite{similar}, where the loss gradient of the last-layer parameters is calculated and used as the vector representation. In both works, the loss for unlabeled instances is computed using the hypothesized label. For labeled instances, the loss is computed using the ground-truth label. 

\subsection{Semi-Hard Selection}

\begin{algorithm}
    \caption{SMI Selection}
    \label{alg:semi_hard}
    \begin{algorithmic}
        \STATE {\bfseries Input:} Labeled set $\mathcal{L}$, Unlabeled set $\mathcal{U}$, Model $M$, Model parameters $\theta$, Budget $b$, Class set $\mathcal{C}$
        \STATE $\hat{G} \leftarrow \{\nabla_{\theta_{last}} L_\theta(x,\hat{y}) | x\in \mathcal{U}, \hat{y} = \text{argmax}_i M_\theta(x)\}$
        \STATE $k \leftarrow \frac{b}{|\mathcal{C}|}$
        \FOR{$c$ \textbf{in} $\mathcal{C}$}
            \STATE $\mathcal{L}_c \leftarrow \{(x,y) \in \mathcal{L} | y = c\}$
            \STATE $G_c \leftarrow \{\nabla_{\theta_{last}} L_\theta(x,y) | (x,y)\in \mathcal{L}_c\}$
            \STATE $\mathcal{S}_{\mathcal{L}_c} \leftarrow \text{CosSimKernel}(G_c, G_c)$
            \STATE $\mathcal{S}_{\mathcal{U},\mathcal{L}_c} \leftarrow \text{CosSimKernel}(\hat{G}, G_c)$
            \STATE $\mathcal{S}_{\mathcal{U}} \leftarrow \text{CosSimKernel}(\hat{G},\hat{G})$
            \STATE $I_f \leftarrow \text{LogDetMI}(\mathcal{S}_{\mathcal{L}_c}, \mathcal{S}_{\mathcal{U},\mathcal{L}_c}, \mathcal{S}_{\mathcal{U}})$
            \STATE $A_c \leftarrow \text{argmax}_{X\subseteq \mathcal{U},|X|\leq k} I_f(X;\mathcal{L}_c)$
        \ENDFOR
        \STATE $A \leftarrow \cup_{c \in \mathcal{C}} A_c$
        \STATE $A_{suggested} \leftarrow \emptyset$
        \FOR{$x$ \textbf{in} $A$}
            \STATE $\hat{c} \leftarrow \text{MaxMarginal}(x,\{A_c|c\in \mathcal{C}\})$
            \STATE $A_{suggested} \leftarrow A_{suggested} \cup \{(x,\hat{c})\}$
        \ENDFOR
        \STATE \textbf{return} $A_{suggested}$
    \end{algorithmic}
\end{algorithm}

We detail our selection of the intermediate unlabeled points in Algorithm~\ref{alg:semi_hard}. As mentioned in Section~\ref{sec:method}, selecting instances of intermediate hardness is achieved on a per-class basis using simple greedy submodular maximization~\cite{nemhauser}. For an intermediate hardness selection budget of $b$, $b/|\mathcal{C}|$ instances are chosen for each class $c$ in $\mathcal{C}$ by maximizing the submodular mutual information (\textsc{LogDetMI} as mentioned previously) between the unlabeled data $\mathcal{U}$ and labeled class exemplars $\mathcal{L}_c$. Notably, all selected samples $A_c$ for class $c$ can be suggested as having a potential label $c$ due to the query relevance with $\mathcal{L}_c$ as mentioned in Section~\ref{sec:method}. However, in scenarios with high class confusion (such as fine-grained classification scenarios such as those presented in Section~\ref{sec:experiments}), there may be instances in $\mathcal{U}$ that get selected more than once (\emph{e.g.}, are present in multiple $A_c$) and thus have multiple class suggestions. To remedy this, we assign such an instance with a suggested label that reflects the $A_c$ to which it contributes the largest marginal gain. In doing so, the instance is assigned the suggested label that most heavily matches the corresponding $\mathcal{L}_c$.

\end{document}